\documentclass[conference]{IEEEtran}
\IEEEoverridecommandlockouts
\usepackage{cite}
\usepackage{balance}
\usepackage{amsmath,amssymb,amsfonts}
\usepackage{graphicx}
\usepackage{textcomp}
\usepackage{xcolor}
\usepackage{amsthm}
\usepackage{url}
\usepackage[ruled, vlined]{algorithm2e}

{}
{}
\newtheorem{definition}{Definition}{}
{}
{}
{}
\newtheorem{problem}{Problem}{}

\usepackage{varwidth,xcolor}
\usepackage{array}
\usepackage{mathrsfs}
\usepackage{multirow}
\usepackage{booktabs}
\usepackage{mdwmath}
\usepackage{mdwtab}
\usepackage{textcomp}
\usepackage{amssymb}
\usepackage{mathtools}
\usepackage{threeparttable} %%for table footnotes
\usepackage{commath}
\usepackage{booktabs}

\usepackage[tight,footnotesize]{subfigure}

\def\BibTeX{{\rm B\kern-.05em{\sc i\kern-.025em b}\kern-.08em
    T\kern-.1667em\lower.7ex\hbox{E}\kern-.125emX}}
\begin{document}
\title{PowerPlanningDL: Reliability-Aware Framework for On-Chip Power Grid Design using Deep Learning}
%\title{PGML: Machine Learning Approach for Fast Electromigration-Aware Aging Prediction of Incremental On-Chip Power Grid Network Design}
%\title{Machine Learning Approach for Fast Electromigration-Aware Aging Prediction of Incremental On-Chip Power Grid Network Design in VLSI SoC}


\author{\IEEEauthorblockN{Sukanta Dey}
\IEEEauthorblockA{\textit{Dept. of CSE,}
\textit{IIT Guwahati}\\
Guwahati, Assam, India \\
sukanta.dey@iitg.ac.in}
\and
\IEEEauthorblockN{Sukumar Nandi}
\IEEEauthorblockA{\textit{Dept. of CSE,} 
\textit{IIT Guwahati}\\
Guwahati, Assam, India \\
sukumar@iitg.ac.in}
\and
\IEEEauthorblockN{Gaurav Trivedi}
\IEEEauthorblockA{\textit{Dept. of EEE,}
\textit{IIT Guwahati}\\
Guwahati, Assam, India \\
trivedi@iitg.ac.in}
}

\maketitle

\begin{abstract}
With the increase in the complexity of chip designs, VLSI physical design has become a time-consuming task, which is an iterative 
design process. Power planning is that part of the floorplanning in VLSI physical design where power grid networks are designed 
in order to provide adequate power to all the underlying functional blocks. Power planning also requires multiple iterative steps 
to create the power grid network while satisfying the allowed worst-case IR drop and Electromigration (EM) margin. For the first 
time, this paper introduces Deep learning (DL)-based framework to approximately predict the initial design of the power grid 
network, considering different reliability constraints. The proposed framework reduces many iterative design steps and speeds 
up the total design cycle. Neural Network-based multi-target regression technique is used to create the DL model. Feature 
extraction is done, and training dataset is generated from the floorplans of some of the power grid designs extracted from 
IBM processor.  The DL model is trained using the generated dataset. The proposed DL-based framework is validated using a 
new set of power grid specifications (obtained by perturbing the designs used in the training phase). The results show that 
the predicted power grid design is closer to the original design with minimal prediction error ($\sim$2\%). The proposed DL-based approach 
also improves the design cycle time with a speedup of $\sim$6$\times$ for standard power grid benchmarks.
\end{abstract}
%With the increase in the complexity of chip designs, VLSI physical design has become a time-consuming task, which is an iterative design process. Power planning is that part of the floorplanning in VLSI physical design where power grid networks are designed in order to provide adequate power to all the functional blocks. Power planning also requires multiple iterative steps to create the power grid network while satisfying the allowed IR drop and Electromigration (EM) margin. For the first time, this paper introduces Deep learning (DL)-based framework to approximately predict the initial design of the power grid network, considering different reliability constraints. The proposed framework reduces many iterative design steps and speeds up the total design cycle. Neural Network-based multi-target regression technique is used to create the DL model. Feature extraction is done, and training dataset is generated from the floorplans of some of the power grid designs extracted from IBM processor.  The DL model is trained using the generated dataset. The proposed DL-based framework is validated using a new set of power grid specifications (obtained by perturbing the designs used in the training phase). The results show that the predicted power grid design is closer to the original design with minimal prediction error. The proposed DL-based approach also improves the design cycle time significantly.


\section{Introduction}\label{sec:introduction}
The primary objective of the power planning phase in the backend-design of a System-on-chip (SoC) is to design a power grid network which can deliver power to 
all the components of the SoC within the allowed margin of IR drop and Electromigration (EM) for the durability of the chip. 
If these margins are not satisfied, then IR drop and EM violation can occur, which reduces the reliability of the chip.
Designing a reliable power grid is an iterative process which requires many phases of incremental design to verify the power grid, as shown in Fig. \ref{fig_pgdesign}. As a 
result of this, the design cost and power planning sign-off time increases. Therefore, to reduce the cost and the design cycle time, in this work we 
propose to utilize the historical data of the power planning design cycle and come up with a deep learning model which can generate a reliable 
power grid. Adaptation of our deep learning model in the power planning phase within the electronics design and automation (EDA) industry reduces 
cost and increases the efficiency of the total design phase of the chip.

\textbf{In this work, we are the first to:}
\begin{itemize}
 \item Present a power planning methodology using the Deep Learning Approach in the VLSI Physical design cycle.
 \item We present a new aspect of obtaining a similarity between power grid design and deep learning. 
 We also build a reliability-aware framework for power grid design using deep learning.
 \item We demonstrate $\sim$6$\times$ speedup in power grid design using the proposed framework compared to the conventional approach for power grid designs of IBM processor.
\end{itemize}
\textbf{At the VLSI Physical Design level, we answer the following questions:}\\
1) How much practically feasible Deep Learning is for the Power Planning phase?\vspace{0.1cm}\\
2) How accurately can the Deep Learning approach predict different design parameters, while still satisfying the allowed IR drop and EM margin?\vspace{0.1cm}\\
3) What is the efficiency of the Deep Learning approach compared to the standard power planning tools?\vspace{0.1cm}\\
These are the fundamental questions that need to be addressed for the successful adaptation of Deep learning approach in the power planning phase.

The paper is arranged as follows. Section \ref{sec_preli} contains all the necessary preliminary details and motivation of the 
manuscript. Section \ref{sec_non_linear} shows the nonlinear formulation of the power grid design problem and its equivalence with
deep learning training, which is used for solving the power grid design problem. Section \ref{sec_proposed_framework} contains
the proposed framework. The experimental results are listed in Section \ref{sec_experimental_result}. The paper is concluded in Section
\ref{sec_con}.
\section{Preliminaries and Motivation}\label{sec_preli}
\subsection{Fundamentals of Power Planning}
\begin{figure}[htbp]
\centering
\includegraphics[scale=0.4]{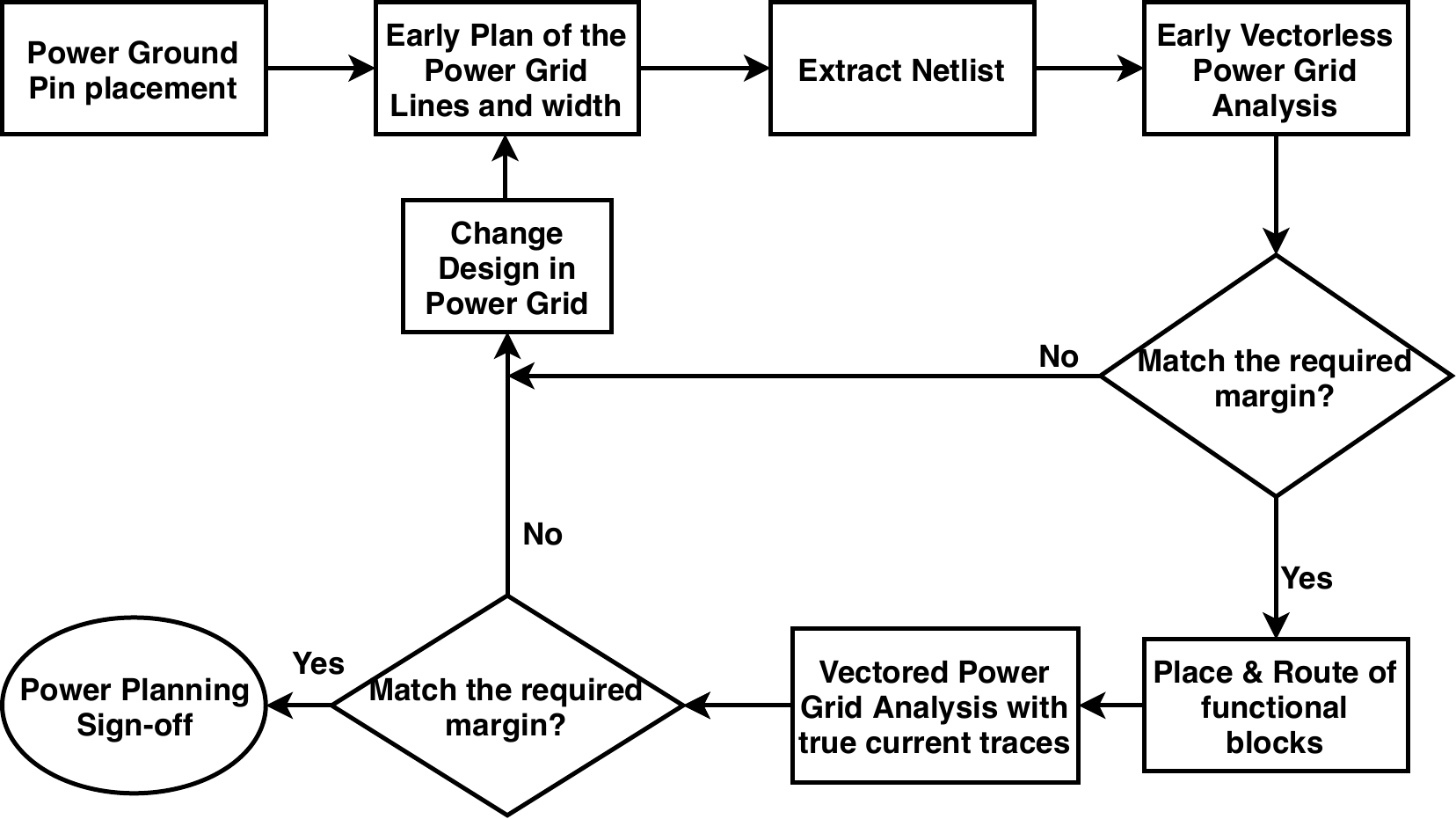}
\caption{Conventional Power Planning Flow in VLSI Physical Design}\label{fig_pgdesign}
\end{figure}
Power Planning is one of the most critical stages in VLSI Physical design. The conventional power planning steps are shown in 
Fig. \ref{fig_pgdesign}. Power planning starts with the pin placement phase of the power and ground pads. Power network is 
generated in order to provide power to standard cells and macros within the acceptable IR-Drop margin. Steady-state IR Drop occurs 
due to the resistance of the metal wires of the power grid network. IR drop can be reduced by decreasing the voltage 
differences between different nodes, which is determined by the power grid analysis. Early vectorless power grid analysis is 
done in order to find the IR drop even before the placement and routing stage with the power information from the front end 
design. Once the margin of IR drop limit is satisfied in this stage, then the placement and routing are done. Subsequently, 
vectored power grid analysis is performed with the exact current traces of the underlying functional blocks in order to 
satisfy the IR drop margin. This work is a first-of-its-kind using a deep learning approach and focuses on the static IR drop and EM-aware power grid design. Therefore, this work does 
not consider the decoupling capacitor (decap) placement phase.
\subsection{Related Work}
\subsubsection{Conventional Approaches in Power Grid Design}
There are many works in the literature in last two decades which deals with power grid designs, analysis, optimization and verification using 
different heuristics. Some of the recent works on the power grids are discussed here.
Fawaz et al. \cite{fawaz2016accurate} have proposed a methodology for accurate verification of the power grids.
Wang et al. \cite{wang2017physics} have proposed electromigration-aware power grid design. Dey et al. \cite{dey2018pgirem} have done 
power grid design considering IR drop and EM reliability constraints. Heo et al. \cite{ik2019detailed} have done IR drop 
mitigation by inserting power staple. All the methods mentioned above suffer from large convergence time.
\subsubsection{Learning Approaches in Power Grid Design}
There are very less efforts for the application of learning-based methods in power grid design. However, few closely related works are
discussed here. Cui et al. \cite{cui2019machine} proposed a machine learning technique for power grid analysis by doing matrix-reordering. 
Fang et al. \cite{fang2018machine} proposed machine learning-based dynamic IR drop prediction.
Liu et al. \cite{liu2016psn} proposed power supply noise aware circuit test timing prediction using machine learning.
Chang et al. \cite{chang2017generating} in their work proposed to generate routability-driven power grid network using machine learning techniques.
Lin et al. \cite{lin2018ir} proposed IR drop prediction of ECO-revised using machine learning.
Ye et al. \cite{ye2014chip} proposed the voltage droop mitigation using support vector deep.
Cao et al. \cite{cao2019learning} proposed a learning-based method to predict the quality of power grid network package.
There is not much significant work in the literature on the deep learning-based power planning methodology.

\subsection{Motivation}
Designing a power grid is similar to solving a non-linear optimization problem, which is proved in the next section. 
Similarly, training deep neural networks is considered as solving a non-linear optimization problem. Therefore, we try to 
investigate the underlying similarity between the two problems and try to solve the power grid design problem
using deep learning. Apart from that, deep learning has been successful in predicting complex tasks in many areas of science 
and technology. Therefore, we use deep learning for prediction of the power grid design, which reduces the design cycle 
time and dependence of human intervention for the initial design of the power grid.
\subsection{Overview of the Proposed Methodology}
\begin{figure}[htbp]
\setlength\abovecaptionskip{-0.3\baselineskip}
\centering
\includegraphics[scale=0.45]{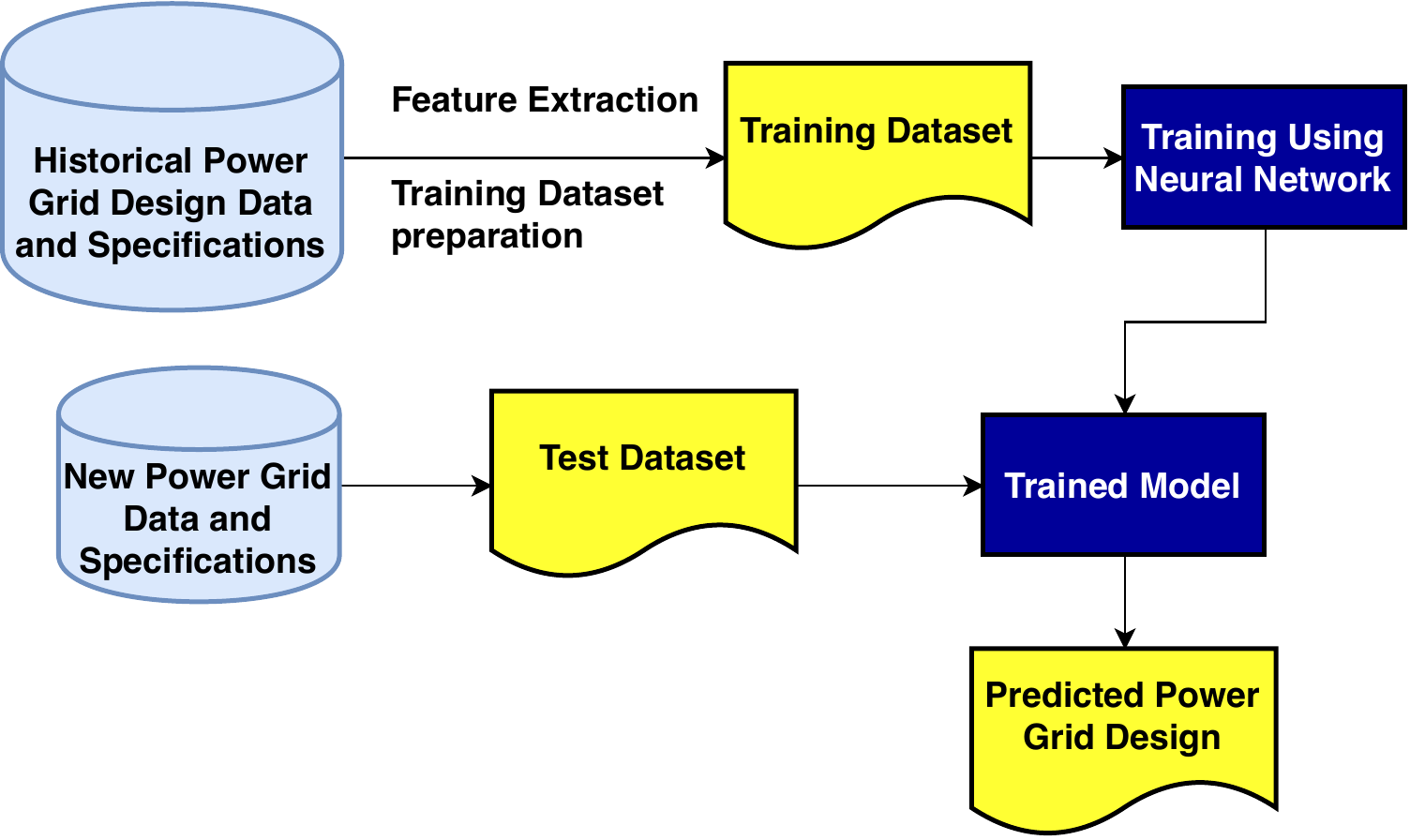}
\caption{Proposed Deep Learning-based Power Planning Flow}\label{fig_pgdesignml}
\end{figure}

Our objective here is to reduce the iterative flow of the power planning phase while still satisfying the allowed margin of IR drop and 
Electromigration with the help of the historical data generated in the design process of the power grid network. Therefore, initially, we perform the 
feature extraction and prepare the training data using these historical data of the design phase and specifications, as shown in Fig. 
\ref{fig_pgdesignml}. Subsequently, we train our deep learning model using these historical data and predict a power grid design for any new design 
specifications.

\section{Nonlinear Optimization Formulation}\label{sec_non_linear}
\begin{figure}[htbp]
\setlength\abovecaptionskip{-0.3\baselineskip}
\includegraphics[trim={0 0 0 0.8cm},clip,scale=0.62]{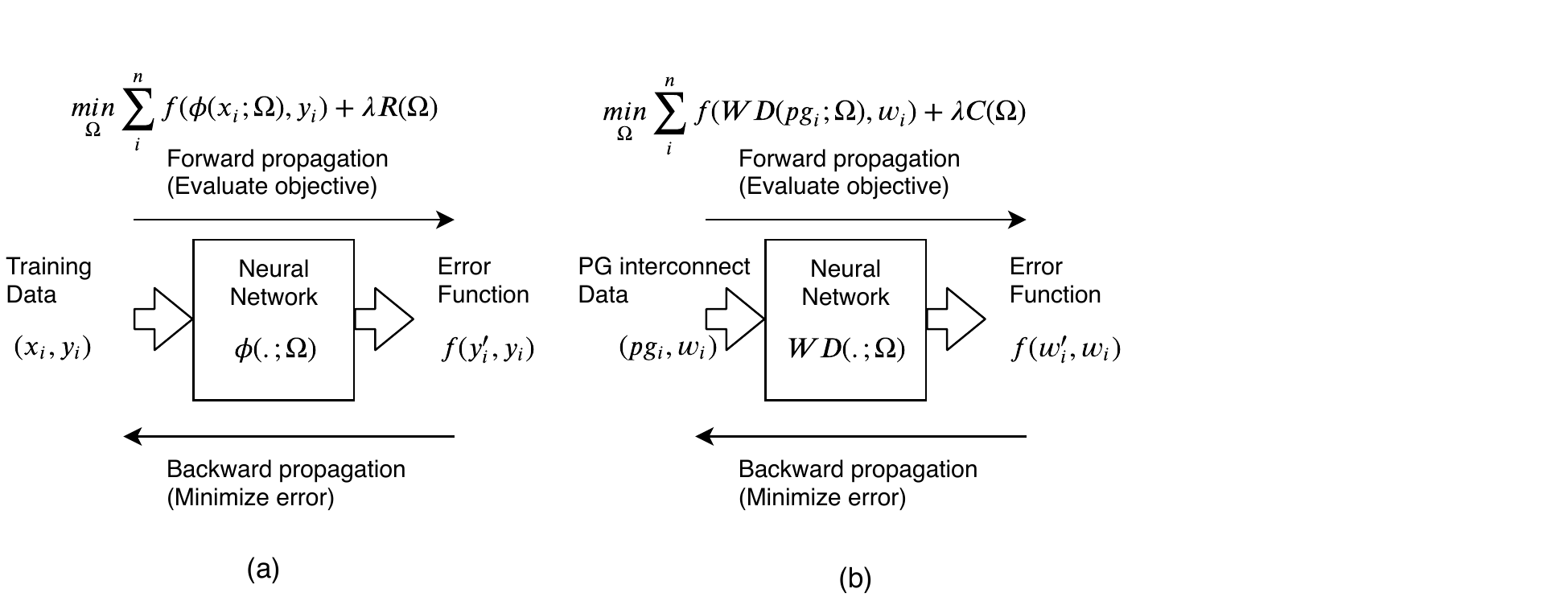}
\caption{Equivalence between deep neural network training and power grid design (a) Training a neural network for weights $\Omega$ 
(b) Solving power grid design with neural networks for weights $\Omega$.}
\label{fig_nn_pg_equivalence}
\end{figure}

In this section, we prove the equivalence between power grid design and deep learning.
The objective of the power grid design is to obtain the optimum width of the power grid lines considering different
reliability constraints.
If the IR drop across the $i^{th}$ power grid lines ($V_{IR_i}$) is represented as
$V_{IR_i} = I_i R_i$,
where $R_i = \rho \frac{l_i}{w_i}$, $\rho$ is sheet resistance, $l_i$ = length, $w_i$ = width, $I_i$ = current throught it.  From here we can write that, 
\begin{equation}\label{eq_w1}
 w_i = \rho \frac{l_i I_i}{V_{IR_i}},
\end{equation}
which is nonlinear function with variables $l_i$ and $V_{IR_i}$ (considering $I_i$ to be constant).
It is also well-known from \cite{dnng} that training of a deep neural network is also a nonlinear optimization problem.
Therefore, both the power grid design and training of a deep neural network are similar.
Using this comparison, we build the neural network model for the power grid design problem which is 
shown in Fig. \ref{fig_nn_pg_equivalence}
Then the minimization of the power grid design objective function can be represented as follows,
\begin{equation}\label{eq_objective}
\underset{\Omega }{min} \sum_{i}^{n} f(WD (pg_i;\Omega ), w_i)) + \lambda C(\Omega),
\end{equation}
where $pg_i$ is the each instance of the power grid interconnect, $WD ()$ is the cost function of \eqref{eq_w1} as predicted by the neural networks for weights $\Omega$.
$f()$ is the error function or loss function to evaluate the error form the true value. $C()$ is the reliability and
other constraints of the power grid design which are described below, and can be satisfied using the weight $\lambda$.

The relation between width of the power grid lines ($w_i$) and the spacing between the two power grid lines ($s_i$) can be represented as follows
\begin{equation}\label{eq_ws}
\sum_{i=1}^{K} s_i + w_i = W_{core}, 
\end{equation}
where $W_{core}$ represents the ring width. For large number of power grid lines, designing power grids with such constraints mentioned in \eqref{eq_w1} and 
\eqref{eq_ws} become difficult and tedious process. The EM reliability constraint for maximum current density $J_{max}$ can be defined as,
\begin{equation}
 \frac{I_i}{w_i} \le J_{max}.
\end{equation}
These constraints need to be satisfied while designing the power grid using neural network, which are denoted as $C()$ in \eqref{eq_objective},
can be adjusted with weight $\lambda$.

\section{Proposed PowerPlanningDL Framework}\label{sec_proposed_framework}
\subsection{Problem Formulation}
\begin{figure}[htbp]
\setlength\abovecaptionskip{-0.3\baselineskip}
\centering
\subfigure[]{\includegraphics[scale=0.24]{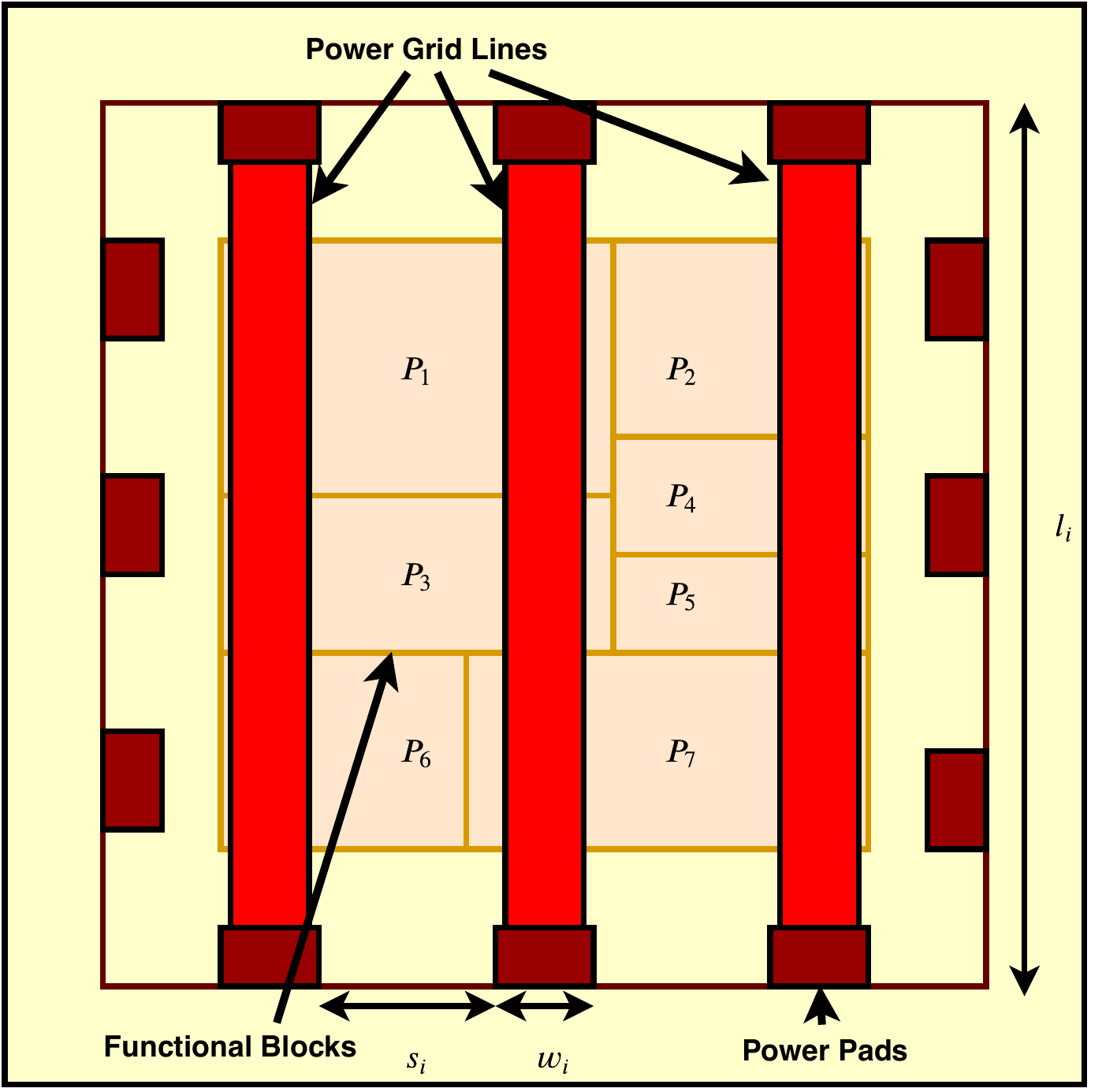}
 \label{fig_problemformulation}}
 \subfigure[]{\includegraphics[scale=0.31]{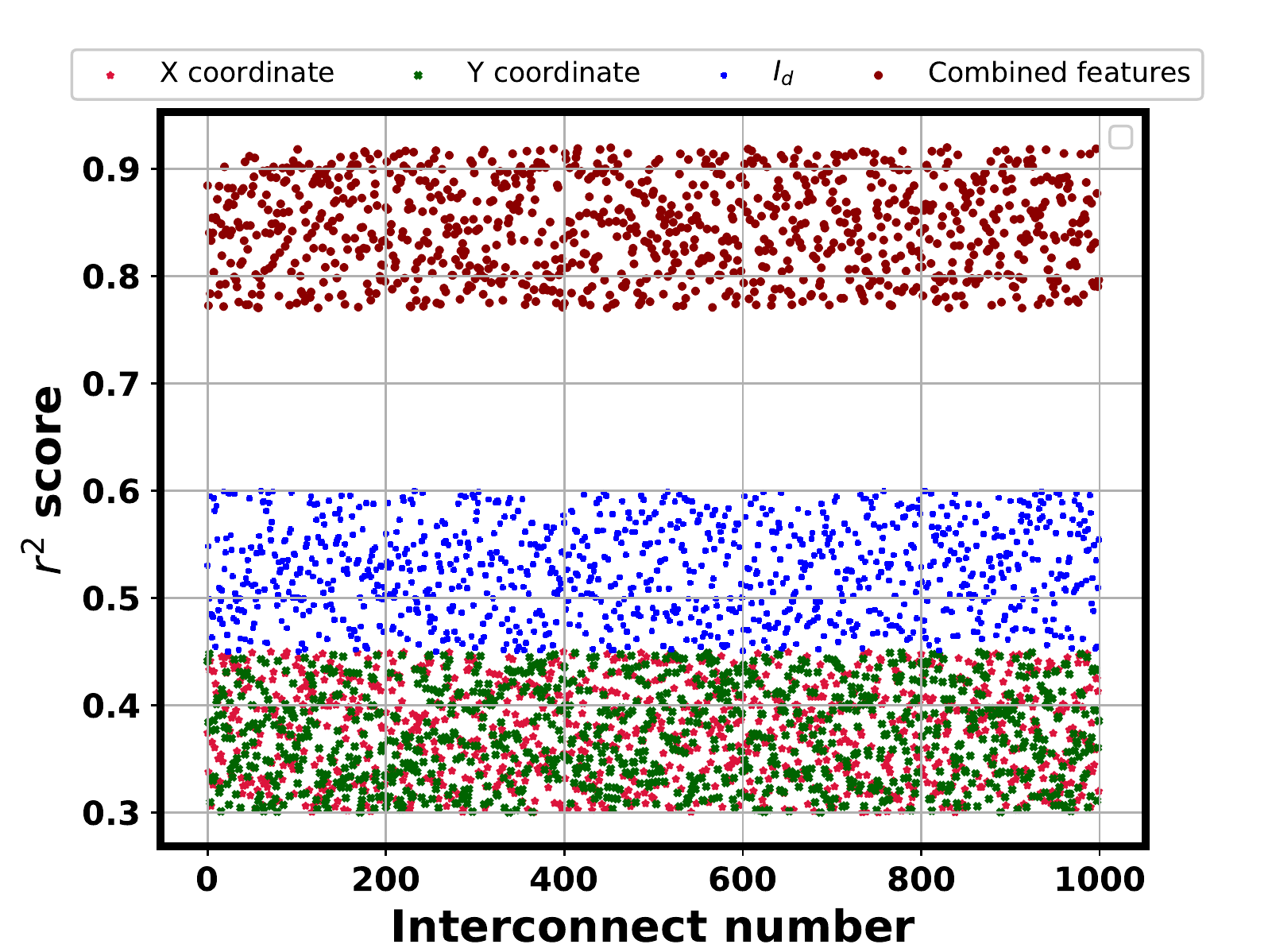}
 \label{fig_r2_v_instance}}
\caption{(a) A floorplan of an SoC with the power grid lines over the functional blocks. (b) Variation of $r^2$ scores for 1000 power grid interconnects of $ibmpg1$ benchmark circuit with different input features}
\label{fig_3}
\end{figure}

A floorplan of an SoC with the power grid lines and underlying functional blocks is shown in Fig. \ref{fig_problemformulation}.
While designing the power grid, it is very challenging to predict the optimum widths of the power grid lines. Overdesigning the power grid lines by increasing the power 
grid line widths increase the total metal routing area of the chip. If it is under-design in order to reduce the metal routing area, then the power 
grid suffers from unwanted IR drop and Electromigration effects due to the increase in resistance and current density of the metal lines. Simultaneously, the 
design rules need to be taken care of while overdesigning/under-designing. The correct predictions of the widths of the power grid lines can reduce 
different iterations of the power planning phase. Therefore, in our deep learning adaptation, we use a supervised learning approach to create a 
model. Our model learns the optimum widths of the metal lines from previous historical data which are obtained for IR drop and Electromigration 
resistant power grid designs with some allowed margin. Subsequently, we use this learned deep learning model in order to predict the widths of 
power grid lines for a new design.

As shown in \eqref{eq_w1}, $w_i$ is dependent on $V_{IR_i}$ and $I_i$, which can only be found after power grid analysis. 
As power grid analysis is time-consuming, we want to evade the power grid analysis phase.
Therefore, we are using alternate approach to predict $w_i$. We are using X-coordinate, Y-coordinate (of the planned floorplan 
of the underlying functional blocks), and its 
switching current activity ($I_d$) (which is obtained from the from front-end phase in value change dump (VCD) file) to predict $w_i$. The reason for choosing these as features are shown in Section \ref{sec_feature}.
Considering this we have formulated two problems to be solved given as follows,
\begin{problem}
Given an X-coordinate, Y-coordinate of floorplan and the switching activity of the current for that point, then predict the metal width 
required for that location which can satisfy the IR drop and EM constraints. 
\end{problem}

\begin{problem}
 Given the width and the switching activity of the PG interconnects, predict the IR drop of the PG interconnect.
\end{problem}

% The number of power grid lines which are required can be obtained using the following formula:
% \begin{equation}
%  \#PG\ line = \frac{W_{core}}{w_i}
% \end{equation}
% As shown in the Fig. \ref{fig_problemformulation}, if we consider that $i^{th}$ power grid line carry, $I_i$ current. Then the 
% current requirement of each of power grid lines to the blocks can be represented as follows,
% \begin{align}
%  I_1 & = I_{11} + I_{13} + I_{16}\\
%  I_2 & = I_{21} + I_{23} + I_{27} \\
%  I_3 & = I_{32} + I_{34} + I_{35} + I_{37},
% \end{align}
% where $I_{ij}$ represents current provided by $i^{th}$ power grid line to the $j^{th}$ block.

We are using a multi-target regression technique to model the deep learning model where we consider multiple input features 
(independent variables, $pg_i$) 
as the input to our model and numerous output features (dependent variables, $w_i$). Mathematically, it can be represented as
\begin{equation}
 \text{Predict } w_i \ \forall i \in \chi,
\end{equation}
where $\chi = \{pg_i\}$ for all $i\in\{1,2,\cdots,n\}$ power grid interconnects is the training dataset.

\subsection{Feature Selection \& Training Data Preparation}\label{sec_feature}
\begin{definition}($r^2$ score) or coefficient of determination is a metric which shows the goodness of the prediction for 
the regression method. A value closer to($\le$) 1 is desired for the data to fit in the model properly. 
\end{definition}

For selecting various features for our deep learning model, we evaluated the $r^2$ score of different input features
with the $w_i$. It has been observed that the combination of the input features X-coordinate, Y-coordinate (of the planned floorplan 
of the underlying functional blocks), and its switching current activity ($I_d$) fits to be the best for the neural network-based
multi-regression technique as it has higher $r^2$ score(Please refer, Fig. \ref{fig_r2_v_instance} and Table \ref{table_r2}).
\begin{table}[htbp]
\centering
\caption{$r^2$ score of different input features and output feature $w$ for a PG interconnect.}\label{table_r2}
\resizebox{3.5in}{!}{%
\begin{tabular}{|l|l|l|l|l|}
\hline
Input Features & X  coordinate & Y  coordinate & $I_d$ & Combined \\ \hline
$r^2$ score & 0.34  & 0.39  & 0.61 & 0.89 \\ \hline
\end{tabular}%
}
\end{table}

$I_d$ is the current obtained from the switching activity of the functional blocks having (X,Y) coordinate. Therefore, the training dataset is generated
with the quadruple (X  coordinate, Y  coordinate, $I_d$, $w_i$) from some of the real power grid desings.
% \begin{figure}[htbp]
% \setlength\abovecaptionskip{-0.3\baselineskip}
% \centering
% \includegraphics[scale=0.4]{./r2_v_instance_number.eps}
% \caption{Variation of $r^2$ scores for 1000 power grid interconnects of $ibmpg1$ benchmark circuit with different input features}\label{fig_r2_v_instance}
% \end{figure}

\subsection{Neural Network-based Deep Learning Model}
\begin{figure}[htbp]
\setlength\abovecaptionskip{-0.3\baselineskip}
\centering
\includegraphics[scale=0.1]{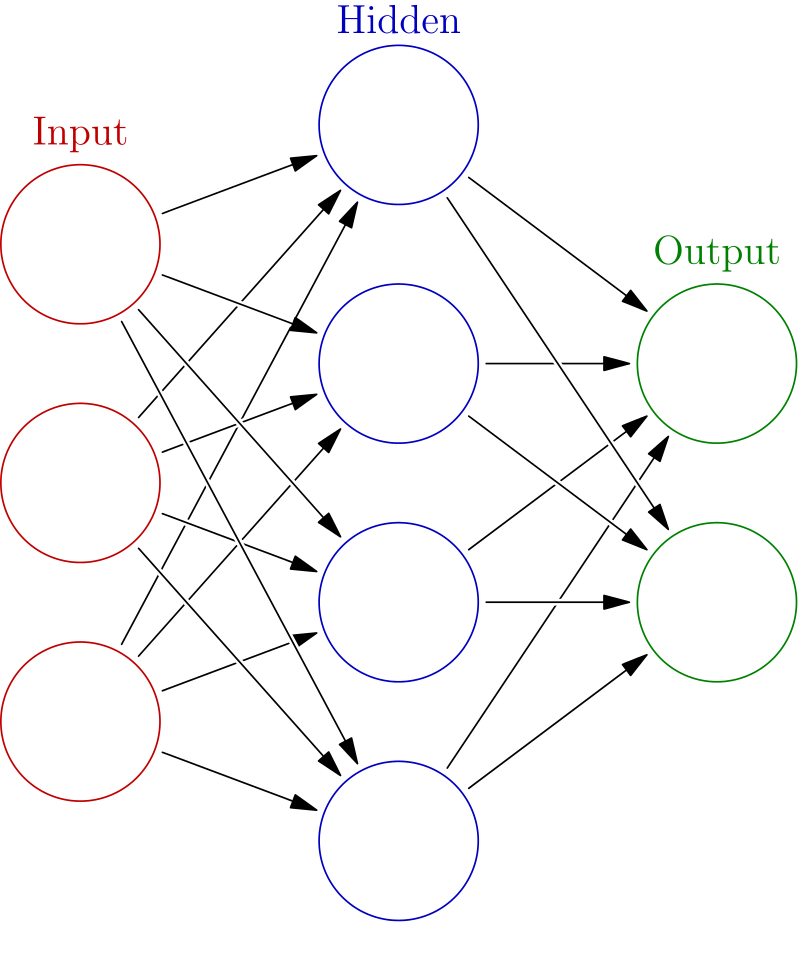}
\caption{Neural Network with one input, one output, and one hidden layer.}
\label{fig_nn}
\end{figure}

The neural network has one input, one output, and hidden layers. An illustrative example is shown in Fig. \ref{fig_nn}.
There can be many number of hidden layers. We have used 10 hidden layers in our model, which is obtained by hyperparameter optimization.
This neural network is trained with quadruple (X  coordinate, Y  coordinate, $I_d$, $w_i$) 
for different weights $\Omega$  as part of its forward propagation step as mentioned in Section \ref{sec_non_linear}. 
Subsequently, adam optimizer \cite{kingma2014adam} is used
to minize the loss or error function as a part in the backpropagation step. Once trained, the new test samples can be 
used to predict $w_i$.
\subsubsection{The Power Grid Interconnect Width Prediction}
The power grid interconnect width prediction is given below in Algorithm \ref{algo_ppdl}.
\begin{algorithm}[htbp]
\setlength\abovecaptionskip{-0.3\baselineskip}
{\fontsize{9}{9}\selectfont
  \KwIn{Training Set}
  %\KwData{this text}
  %\KwResult{how to write algorithm with \LaTeX2e }
 \KwOut{$w_i$ and gradient}
  \everypar={\nl}
  ForwardPropagation(X  coordinate, Y  coordinate, $I_d$, $w_i$)\\
  \{\\
   \hspace{0.5cm} return loss function $f$
    \\ \} \\
  BackwardPropagation()\\
   \{\\
   \hspace{0.5cm} return gradient
    \\ \} \\
  \caption{Wire width prediction by NN}
  \label{algo_ppdl}
}
\end{algorithm}

\subsubsection{IR Drop Prediction}
The IR drop prediction algorithm is given below in Algorithm \ref{algo_ir}.
\begin{algorithm}[htbp]
\setlength\abovecaptionskip{-0.3\baselineskip}
{\fontsize{9}{9}\selectfont
  \KwIn{Predicted width $w_i$}
  %\KwData{this text}
  %\KwResult{how to write algorithm with \LaTeX2e }
 \KwOut{Predicted IR drop}
  \everypar={\nl}
  From switching current $I_d$ and $w_i$\;
  Use kirchoff's law to predict IR drop.
  \caption{IR drop prediction}
  \label{algo_ir}
}
\end{algorithm}
From Algorithm \ref{algo_ppdl} after testing on test dataset, we already have the $w_i$, which means we have the $R_i$ of the power
grid interconnect (considering $l_i$ to be constant). We need the $I_i$ to find the IR drop across the interconnect. The following
approach helps in obtaining $I_i$.
The number of power grid lines which are required can be obtained using the following formula:
\begin{equation}
 \#PG\ line = \frac{W_{core}}{w_i}
\end{equation}
As shown in the Fig. \ref{fig_problemformulation}, if we consider that $i^{th}$ power grid line carry, $I_i$ current. Then the 
current requirement of each of power grid lines to the blocks can be represented as follows,
\begin{align}
 I_1 & = I_{11} + I_{13} + I_{16}\\
 I_2 & = I_{21} + I_{23} + I_{27} \\
 I_3 & = I_{32} + I_{34} + I_{35} + I_{37},
\end{align}
where $I_{ij}$ represents current provided by $i^{th}$ power grid line to the $j^{th}$ block.
From the above, we can obtain current through the interconnect and subsequently the IR drop.
\subsection{Test Data Generation}
Test dataset is generated by perturbing the same dateset which are used for training. The perturbation is done by changing the
branch current, node voltage, and switching current of the underlying functional blocks by a $\gamma = 10\%$, which is termed as
perturbation size. Experiments are done in the next section by varying the perturbation size in order to see the 
variation in prediction accuracy.

\section{Experimental Results}\label{sec_experimental_result}
\subsection{Simulation Setup}
\begin{figure}[htbp]
\setlength\abovecaptionskip{-0.3\baselineskip}
\centering
\includegraphics[scale=0.3]{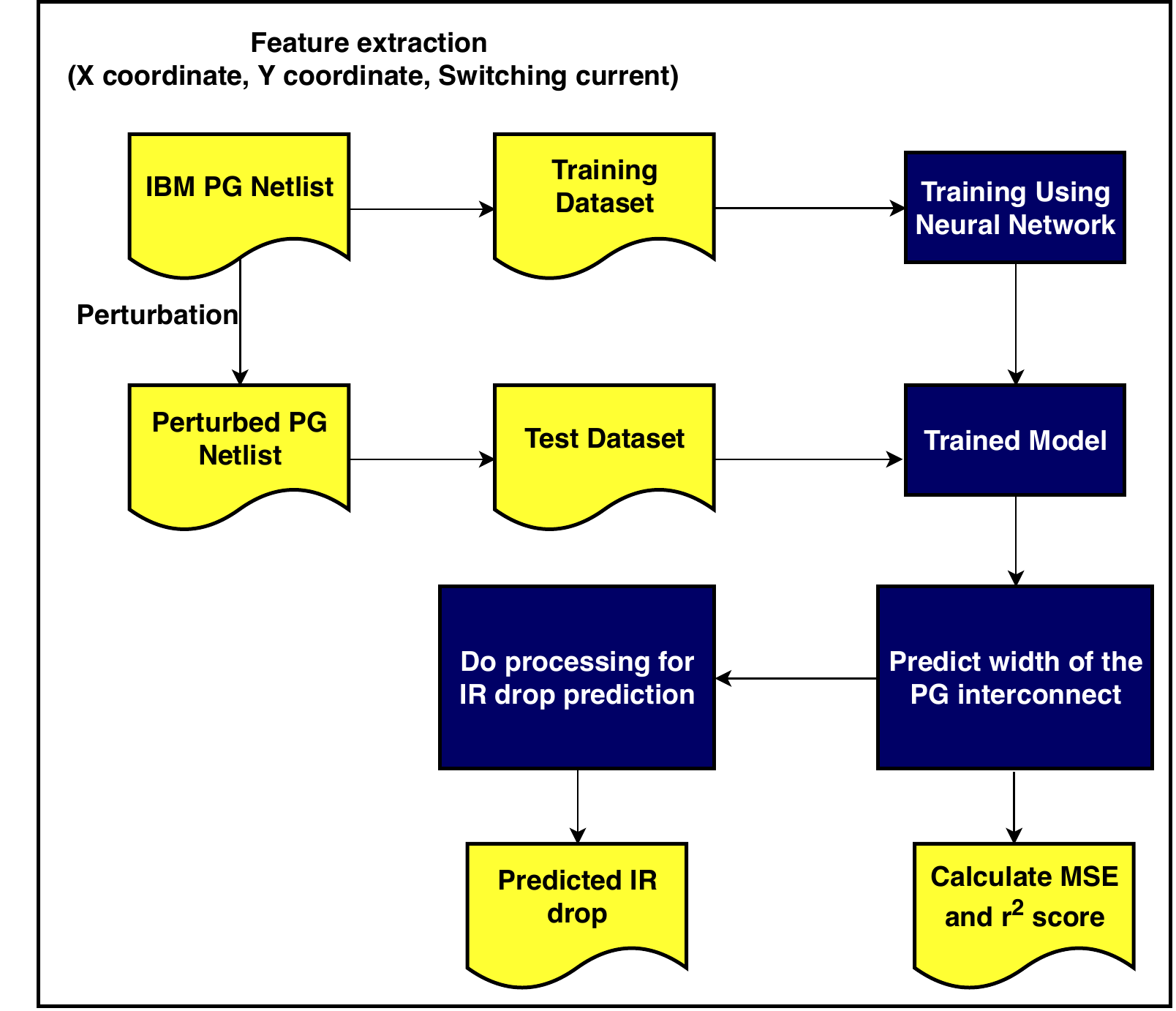}
\caption{Flow of the simulation setup of the Deep Learning Flow}
\label{fig_simulation_setup}
\end{figure}

The framework is developed with \emph{C++} and \emph{python}. For deep learning operations \emph{Tensorflow} library of the 
python has been used on a Linux machine with Intel Xeon E5-2650 processor, with the GPU configuration Nvidia Tesla K20c. 
The datasets are generated, and the proposed PowerPlanningDL is validated using the IBM Power Grid benchmarks 
\cite{nassif2008power}, which are standard power grid benchmarks extracted from IBM processors. The details of the IBM 
PG benchmarks are listed in Table \ref{table_pgbench}. Current loads of the IBM PG benchmarks are modified in order to 
obtain the desired effects.
The simulation setup for the experiments is set according to 
Fig. \ref{fig_simulation_setup}. All the hyperparameters of the neural network are fixed for which the best results are obtained.
\begin{table}[htbp]
\centering
\caption{IBM PG benchmark details \cite{nassif2008power}}
\scalebox{0.8}{
\begin{threeparttable}
\begin{tabular}{|l|l|l|l|l|}
\hline
\textbf{PG Circuits} & \textbf{\#n}     & \textbf{\#r}     & \textbf{\#v}    & \textbf{\#i}    \\ \hline
$ibmpg1$    & 30638   & 30027   & 14308  & 10774  \\ \hline
$ibmpg2$    & 127238  & 208325  & 330    & 37926  \\ \hline
$ibmpg3$    & 851584  & 1401572 & 955    & 201054 \\ \hline
$ibmpg4$    & 953583  & 1560645 & 962    & 276976 \\ \hline
$ibmpg5$    & 1079310 & 1076848 & 539087 & 540800 \\ \hline
$ibmpg6$    & 1670494 & 1649002 & 836239 & 761484 \\ \hline
$ibmpgnew1$ & 1461036 & 2352355 & 955    & 357930 \\ \hline
$ibmpgnew2$ & 1461039 & 1422830 & 930216 & 357930 \\ \hline
\end{tabular}
\begin{tablenotes}[para,flushleft]
 \item \textbf{\#n} : total number of nodes of PG network, 
 \textbf{\#r} : total number of resistors (edges) of PG network, 
 \textbf{\#v} : total number of supply voltage ($V_{dd}$ and $GND$ sources) of PG network, 
 \textbf{\#i} : total number of workloads connected to PG network.
\end{tablenotes}
\end{threeparttable}}\label{table_pgbench}
\end{table}

\subsection{Study of Predicted Power Grid Interconnect Width}
\begin{figure}[htbp]
\setlength\abovecaptionskip{-0.3\baselineskip}
\centering
\subfigure[]{\includegraphics[scale=0.25]{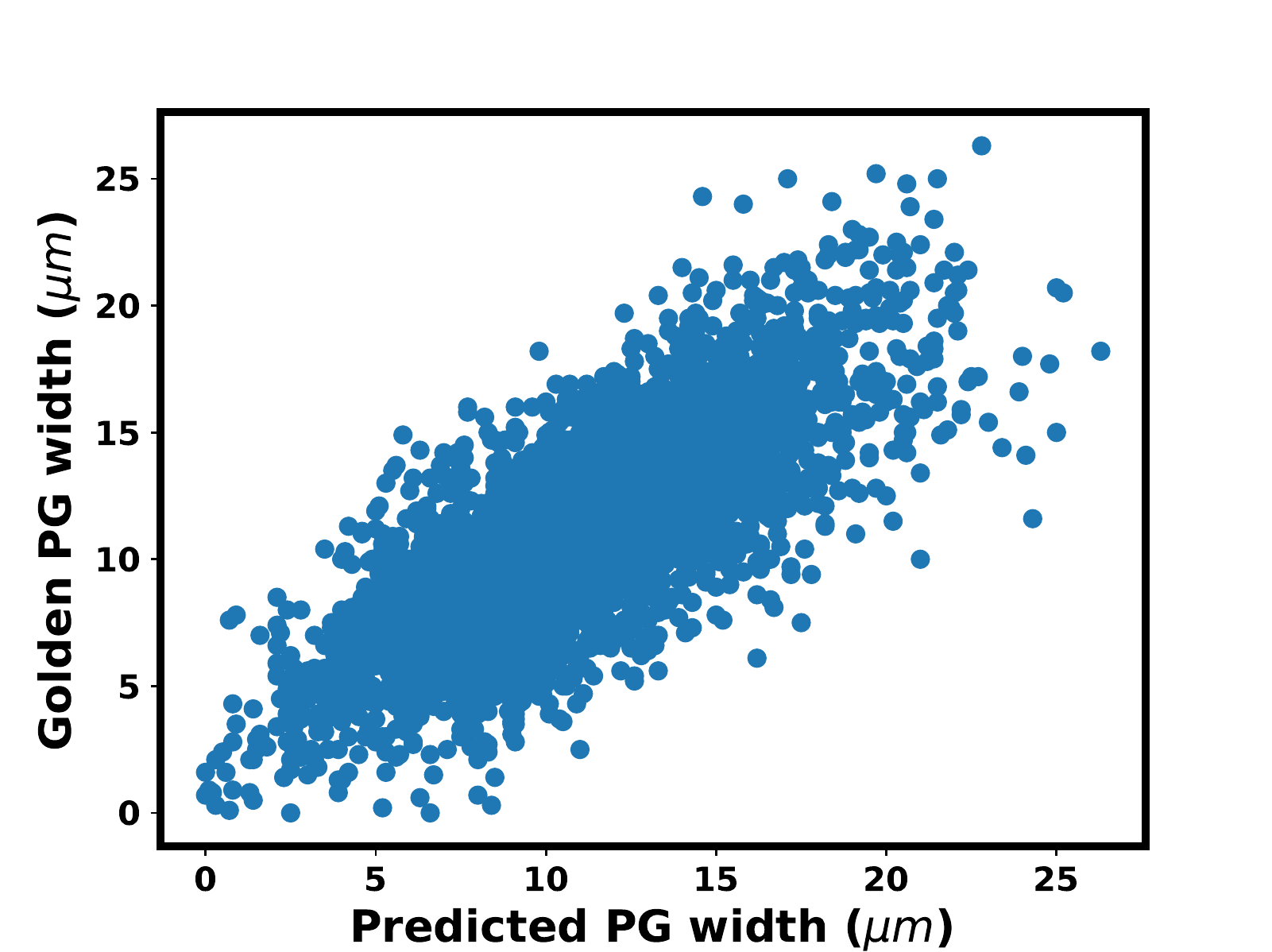}
 \label{fig_scatter_plot}}
 \hfil
 \subfigure[]{\includegraphics[scale=0.25]{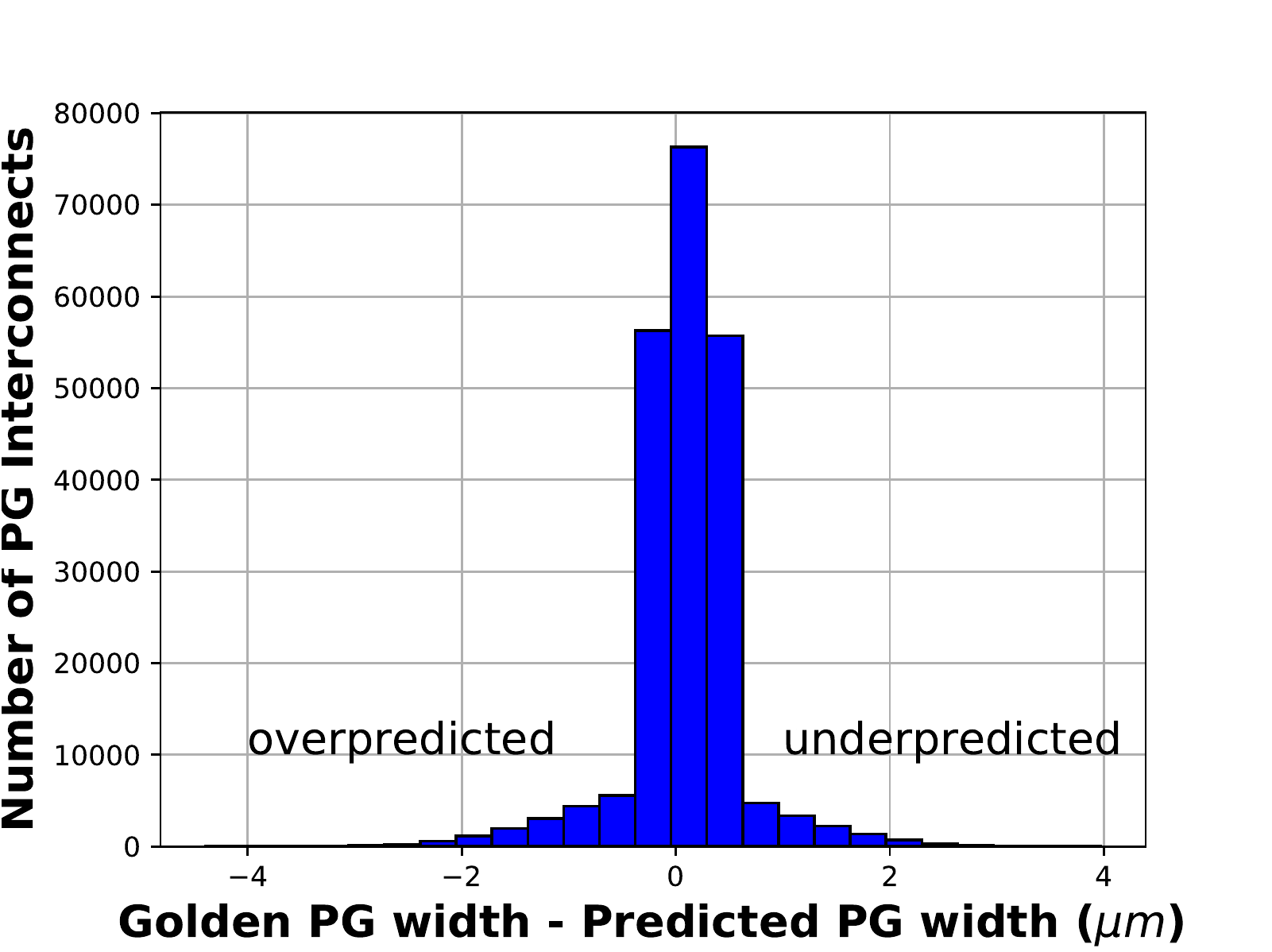}
 \label{fig_histogram}}
\caption{Power Grid interconnect width prediction for $ibmpg2$ benchmark circuit (a) Correlation scatter plot (b) Error histogram (Horizontal axis represent the error).}
\label{fig_wire_perturbation}
\end{figure}

In this section, the correlation between the predicted width of the power grid using PowerPlanningDL and conventional approach 
evaluated. From the correlation value, it can be seen how much the predicted widths are related to the golden width 
obtained from the conventional approach. The correlation plot is shown in Fig. \ref{fig_scatter_plot}. 
To study the error distribution of the predicted widths, the error histogram plot
is shown in Fig. \ref{fig_histogram} (Horizontal axis represent error). From the error histogram, we can observe that 
most of the predicted widths are concentrated near 0, meaning most of the predicted widths of PG interconnect produce near about 0 error. 
As the amount of error increases, the number of power grid instances decreases. From this result, 
we can conclude that the predicted widths of the power grid lines using PowerPlanningDL are very close 
to the golden results generated by the conventional approach for most of the interconnects.

\subsection{Study of Predicted IR Drop in Power Grid}
\begin{figure}[htbp]
\setlength\abovecaptionskip{-0.3\baselineskip}
\centering
\subfigure[]{\includegraphics[scale=0.25]{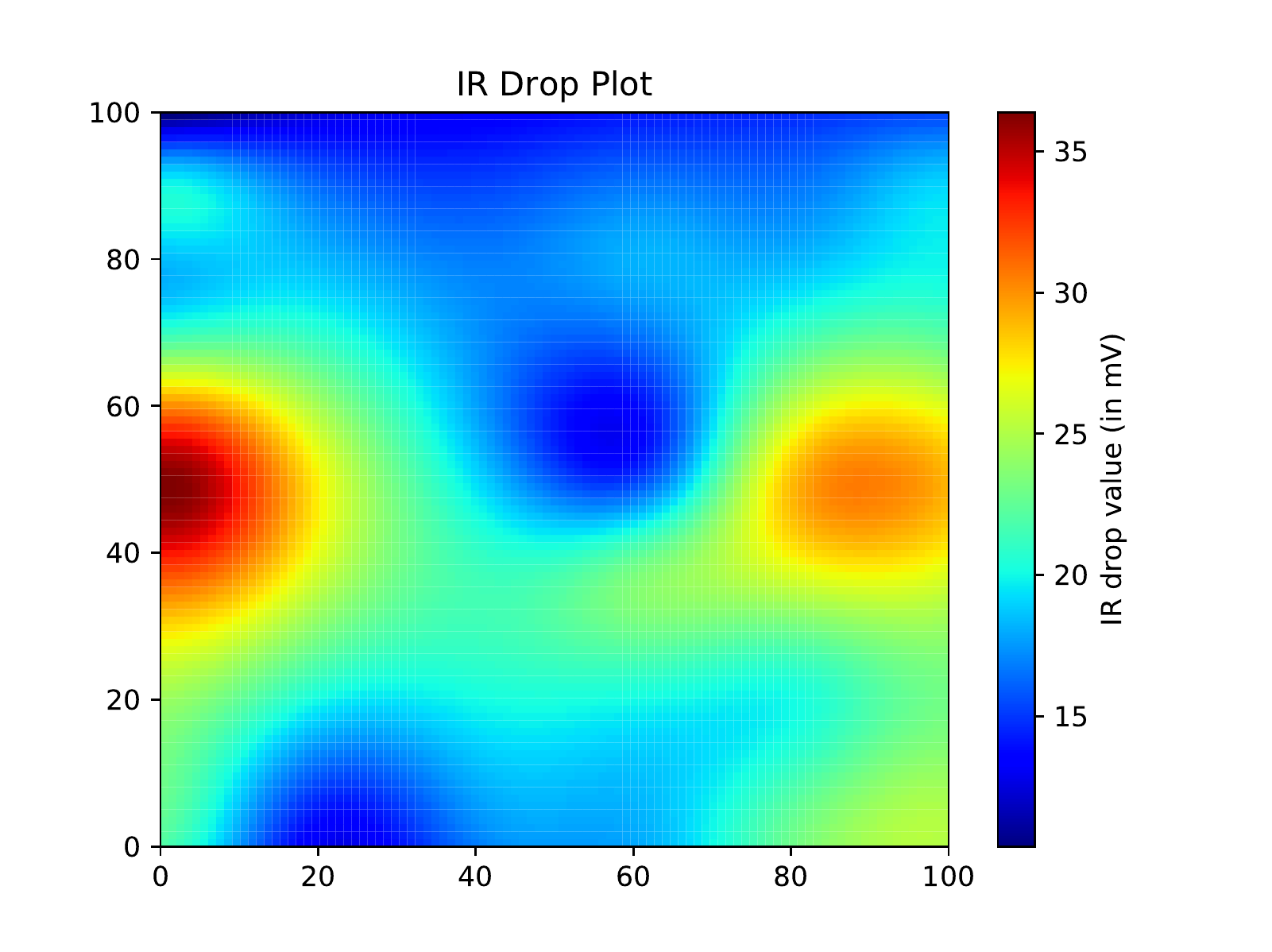}
 \label{fig_ir_c}}
 \hfil
 \subfigure[]{\includegraphics[scale=0.25]{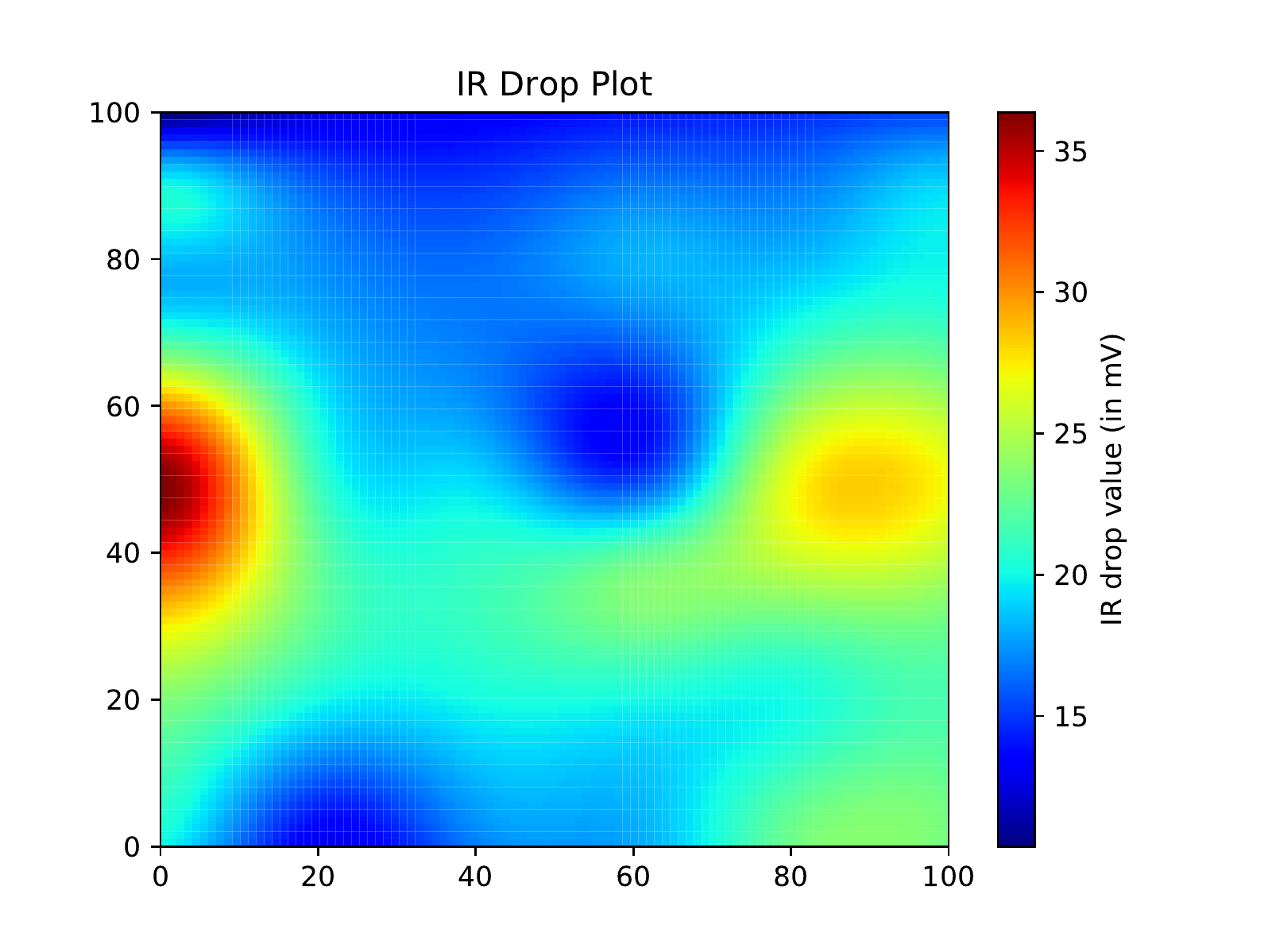}
 \label{fig_ir_dl}}
  \hfil
 \subfigure[]{\includegraphics[scale=0.25]{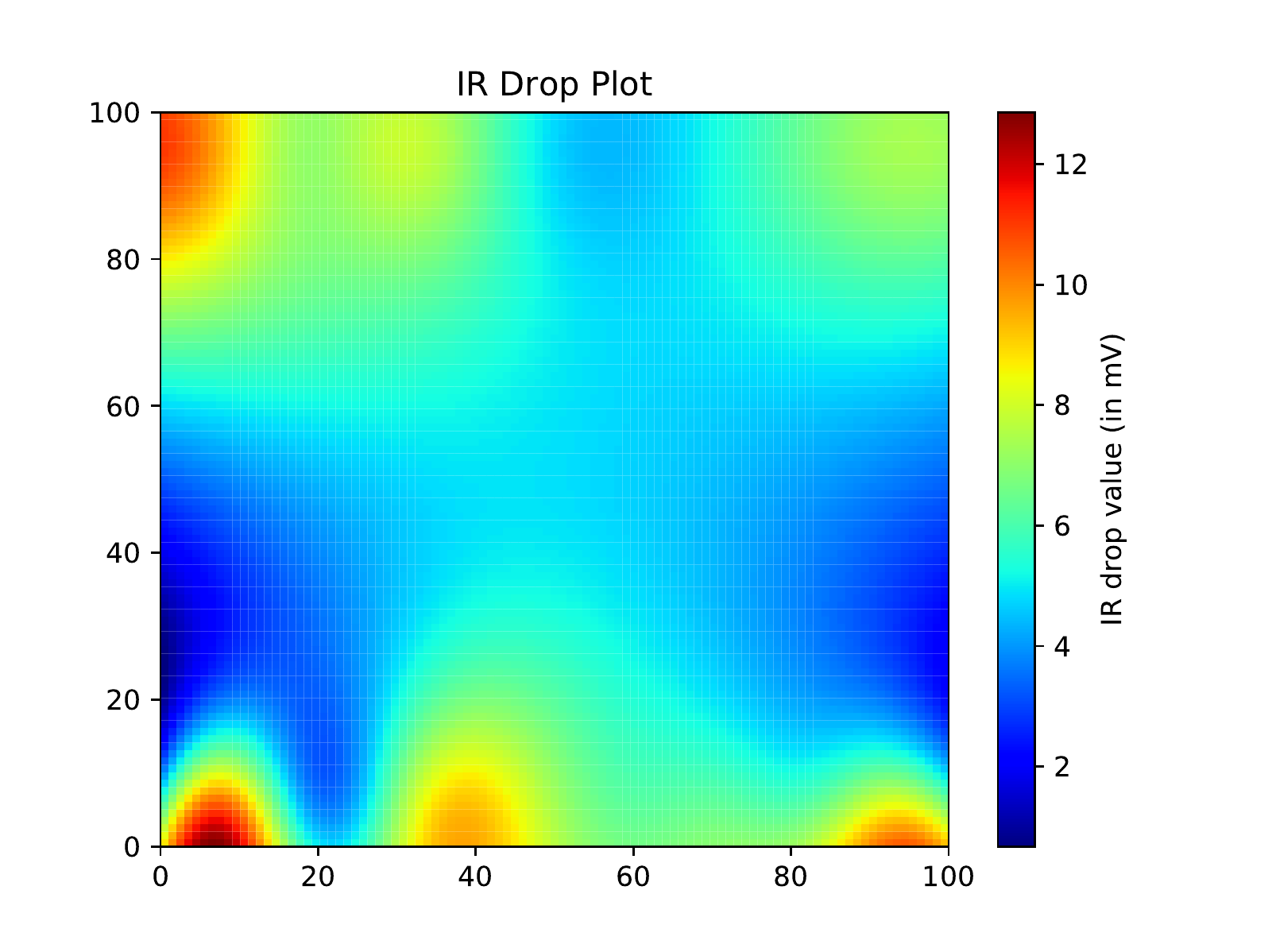}
 \label{fig_ir_dl}}
  \hfil
 \subfigure[]{\includegraphics[scale=0.25]{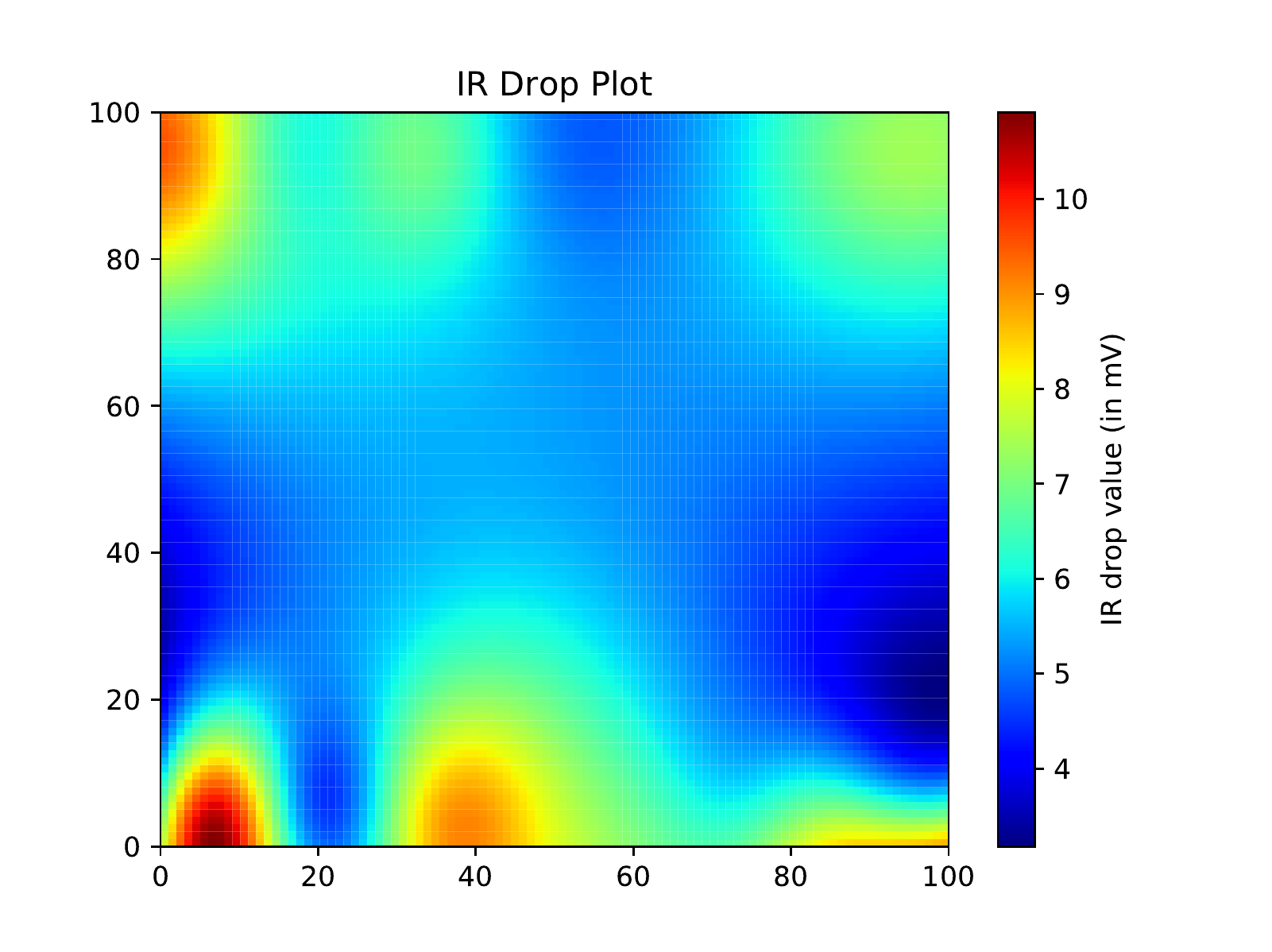}
 \label{fig_ir_dl}}
\caption{IR drop map of (a) Conventional method $ibmpg2$ circuit (b) PowerPlanningDL methodology $ibmpg2$ 
circuit (c) Conventional method $ibmpg6$ circuit, and (d) PowerPlanningDL methodology $ibmpg6$ circuit.}
\label{fig_ir_drop_map}
\end{figure}

The IR drop map is plotted for the conventional approach and also for the PowerPlanningDL approach, as shown in Fig. \ref{fig_ir_drop_map}
for $ibmpg2$ circuit and $ibmpg6$ circuit. The worst-case IR drop for all the benchmarks are listed in 
Table \ref{IRdropanatime}. From the IR drop map and the worst-case IR drop values, it can be inferred that the PowerPlanningDL can predict the IR drop close
to the conventional approach.

\begin{table}[htbp]
\centering 
\caption{Comparision of Worst-case IR drop using Conventional power planning approach and PowerPlanningDL framework}
\label{IRdropanatime}
\scalebox{0.8}{
\begin{tabular}{lll|}
\hline
\multicolumn{1}{|l|}{}             & \multicolumn{2}{l|}{\textbf{Worst-case IR drop ($mV$)}}                              \\ \hline
\multicolumn{1}{|l|}{\textbf{PG circuits}} & \multicolumn{1}{l|}{\textbf{Conventional}} & \multicolumn{1}{l|}{\textbf{PowerPlanningDL}} \\ \hline
\multicolumn{1}{|l|}{$ibmpg1$}       & \multicolumn{1}{l|}{69.8}       & \multicolumn{1}{l|}{68.2}       \\ \hline
\multicolumn{1}{|l|}{$ibmpg2$}       & \multicolumn{1}{l|}{36.3}       & \multicolumn{1}{l|}{36.1}       \\ \hline
\multicolumn{1}{|l|}{$ibmpg3$}       & \multicolumn{1}{l|}{18.1}       & \multicolumn{1}{l|}{18.0}        \\ \hline
\multicolumn{1}{|l|}{$ibmpg4$}       & \multicolumn{1}{l|}{4.0}       & \multicolumn{1}{l|}{4.1}         \\ \hline
\multicolumn{1}{|l|}{$ibmpg5$}       & \multicolumn{1}{l|}{4.3}       & \multicolumn{1}{l|}{4.2}       \\ \hline
\multicolumn{1}{|l|}{$ibmpg6$}       & \multicolumn{1}{l|}{13.1}       & \multicolumn{1}{l|}{13.0}         \\ \hline 
\end{tabular}}
\end{table}

\subsection{Main Result: Study of Convergence Time}
The convergence time for both the approach is shown in Table \ref{convergence_time}. 
Convergence time of the conventional approach includes the IR drop analysis time, as it is the primary time-consuming task.
For the PowerPlanningDL, the convergence time shows the prediction time of the width and IR drop prediction time, as mentioned in Section \ref{sec_proposed_framework}.
From the table, it can be seen that our 
proposed PowerPlanningDL is 5.87$\times$ faster than the conventional approach for the $ibmpg5$ benchmark. It is also
observed that for larger benchmarks the speedup is more, as larger grids take more time for power grid analysis in conventional
approach, which is not used in our PowerPlanningDL framework. That is one of the main reason that we get a significant speedup
for our PowerPlanningDL compared to the conventional approach. We achieve the speedup at the cost of accuracy.
It is to be noted that for the convergence time of the conventional approach reported in Table \ref{convergence_time}, we have 
considered the best-case scenario and reported the convergence time only for one iteration of the design cycle. 
In the worst case, there can be multiple iterations of the design cycle, for which the conventional approach 
takes much more time, whereas the convergence time will be the same for PowerPlanningDL in all scenarios.
This also shows the advantage of PowerPlanningDL in reducing the number of iterations in the design cycle.
\begin{table}[htbp]
\centering 
\caption{Comparison of convergence time for Conventional power planning approach and PowerPlanningDL framework}
\label{convergence_time}
\scalebox{0.8}{
\begin{tabular}{llll|}
\hline
\multicolumn{1}{|l|}{}             & \multicolumn{2}{l|}{\textbf{Time (sec)}}                               & \textbf{Speedup} \\ \hline
\multicolumn{1}{|l|}{\textbf{PG circuits}} & \multicolumn{1}{l|}{\textbf{Conventional}} & \multicolumn{1}{l|}{\textbf{PowerPlanningDL}} & \textbf{$\frac{\text{Time}_{\mathbf{Conventional}}}{\text{Time}_{\mathbf{PowerPlanningDL}}}$}		\\ \hline
\multicolumn{1}{|l|}{$ibmpg1$}       & \multicolumn{1}{l|}{6.85}       & \multicolumn{1}{l|}{3.56}       & 1.92$\times$   \\ \hline
\multicolumn{1}{|l|}{$ibmpg2$}       & \multicolumn{1}{l|}{23.46}       & \multicolumn{1}{l|}{11.88}     & 1.97$\times$     \\ \hline
\multicolumn{1}{|l|}{$ibmpg3$}       & \multicolumn{1}{l|}{29.50}       & \multicolumn{1}{l|}{8.07}      & 3.59$\times$    \\ \hline
\multicolumn{1}{|l|}{$ibmpg4$}       & \multicolumn{1}{l|}{52.4}       & \multicolumn{1}{l|}{11.83}     & 4.42$\times$     \\ \hline
\multicolumn{1}{|l|}{$ibmpg5$}       & \multicolumn{1}{l|}{74.80}       & \multicolumn{1}{l|}{12.74}         & 5.87$\times$  \\ \hline
\multicolumn{1}{|l|}{$ibmpg6$}       & \multicolumn{1}{l|}{97.5}       & \multicolumn{1}{l|}{17.41}     &  5.60$\times$    \\ \hline
\multicolumn{1}{|l|}{$ibmpgnew1$}   & \multicolumn{1}{l|}{102.58}       & \multicolumn{1}{l|}{21.50}     & 4.77$\times$     \\ \hline
\multicolumn{1}{|l|}{$ibmpgnew2$}   & \multicolumn{1}{l|}{48.60}       & \multicolumn{1}{l|}{10.86}      & 4.47$\times$    \\ \hline 
\end{tabular}}
\end{table}

\subsection{Overhead: Study of Model Accuracy}
The mean square error (MSE) can be defined as, 
\begin{equation}
 MSE = \frac{1}{n} \sum_{i=1}^{n} (y_i - y_i')^2
\end{equation}
The $r^2$ score, and MSE using the proposed framework is listed in Table \ref{table_mse}.
MSE tells about the prediction error (overhead of deep learning approach) while predicting the interconnect width.
From this result of MSE, we can conclude that the proposed PowerPlanningDL can predict the power grid design, which is very close to the golden design generated by the conventional approach. 
From $r^2$ score we know how well the data is fit in the model.
\begin{table}[htbp]
\centering
\caption{$r^2$ score, MSE and Peak memory using PowerPlanningDL framework for all the IBM PG benchmarks}\label{table_mse}
%\resizebox{5.5in}{!}{
\scalebox{1}{
% \usepackage{graphicx}
\resizebox{3in}{!}{%
\begin{tabular}{|l|l|l|l|l|}
\hline
\textbf{PG Circuits} & \textbf{\#interconnects} & $\mathbf{r^2}$ \textbf{score} & \textbf{MSE} & \textbf{Peak Memory (in MiB)} \\ \hline
$ibmpg1$ & 30027 & 0.933 & 0.0231  &  66\\ \hline
$ibmpg2$ & 208325 & 0.937  & 0.0230  & 318  \\ \hline
$ibmpg3$ & 1401572 & 0.932   & 0.0212  & 730 \\ \hline
$ibmpg4$ & 1560645 & 0.941  & 0.0210  & 749 \\ \hline
$ibmpg5$ & 1076848 & 0.944  & 0.0225 & 511 \\ \hline
$ibmpg6$ & 1649002 & 0.945  & 0.0208 & 841  \\ \hline
$ibmpgnew1$ & 2352355 & 0.943   & 0.0201  & 1025 \\ \hline
$ibmpgnew2$ & 1422830 & 0.945  & 0.0209 & 745 \\ \hline
\end{tabular}%
}}
\end{table}

\subsection{Study of Variation of MSE with Perturbation Size}
\begin{figure}[htbp]
\setlength\abovecaptionskip{-0.3\baselineskip}
\centering
\subfigure[]{\includegraphics[scale=0.25]{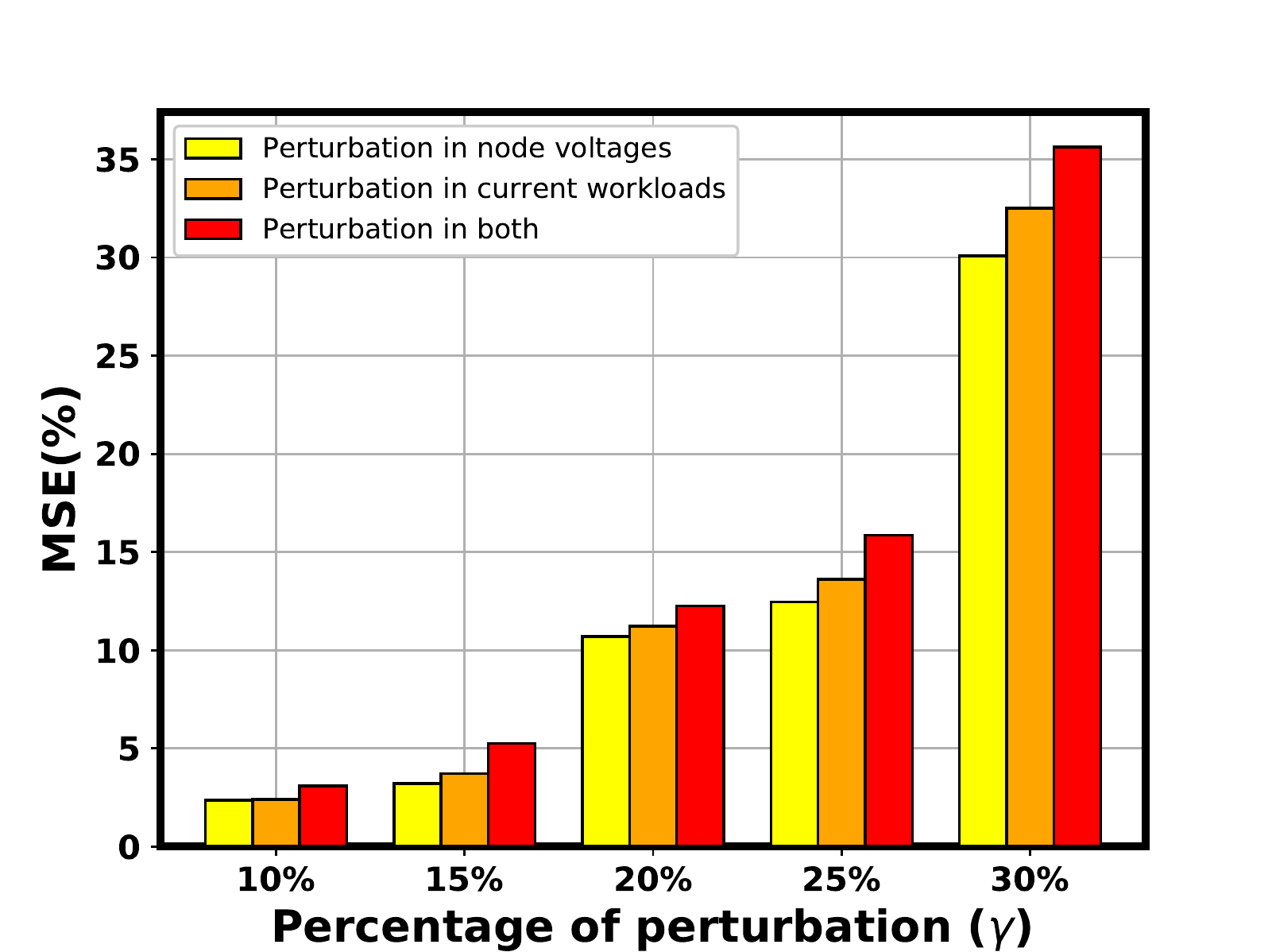}
 \label{fig_mse_perturbation_ibmpg2}}
 \hfil
 \subfigure[]{\includegraphics[scale=0.25]{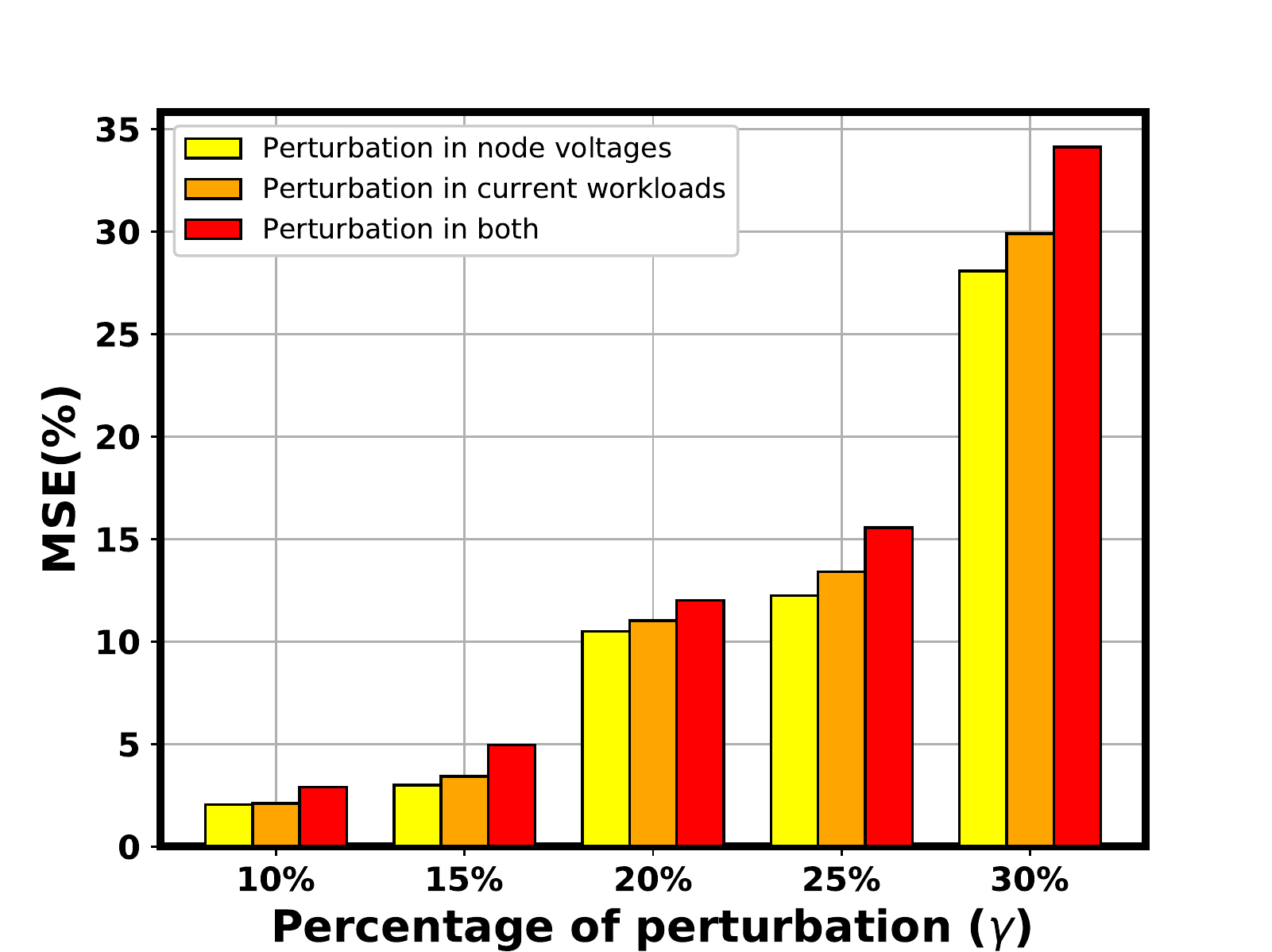}
 \label{fig_mse_perturbation_ibmpg4}}
\caption{Comparison of prediction accuracy on test set in MSE with variations in perturbations size for (a) $ibmpg2$ (b) $ibmpg6$ benchmark circuit.}
\label{fig_mse_perturbation}
\end{figure}

The variation of MSE with the perturbation size ($\gamma \%$) is shown in Fig. \ref{fig_mse_perturbation}. It is observed that as the perturbation
size increases the MSE increases.
From this observation, we can infer that the proposed PowerPlanningDL is best suited for the incremental-based power grid design, where we need to generate the power grid for little changes (or perturbations) in the design.

\subsection{Study of Peak Memory}
\begin{figure}[htbp]
\setlength\abovecaptionskip{-0.3\baselineskip}
\centering
\subfigure[]{\includegraphics[scale=0.2]{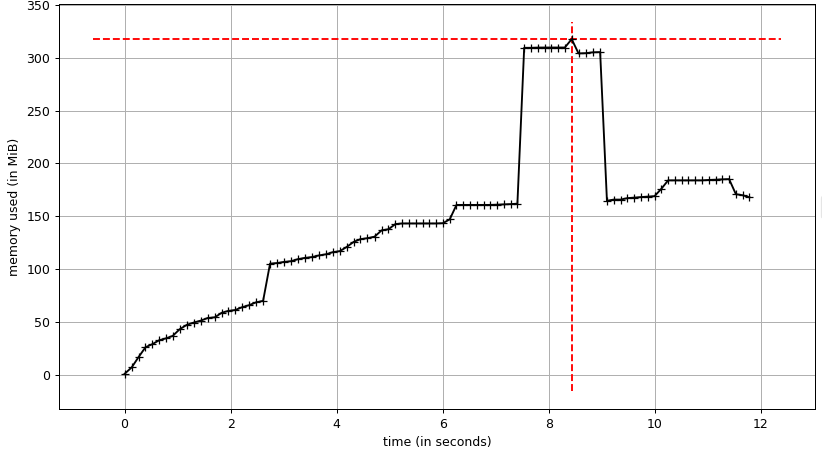}
 \label{fig_peak_memeory_GPR}}
 \hfil
 \subfigure[]{\includegraphics[scale=0.2]{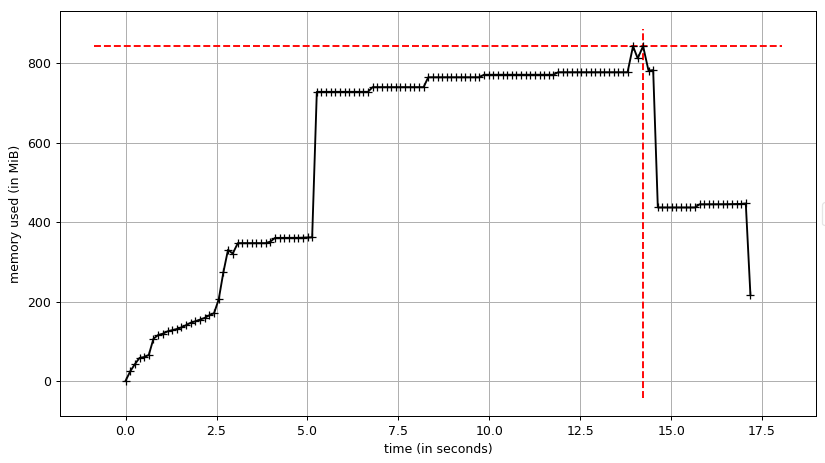}
 \label{fig_peak_memeory_NN}}
\caption{Memory used by PowerPlanningDL for (a) $ibmpg2$ benchmark circuit and (b) $ibmpg6$ benchmark circuit. 
1 Gigabyte (GB) = 953.674 Mebibyte (MiB)}
\label{fig_peak_memeory}
\end{figure}

For the completeness of the results, we have also evaluated the memory profile of the proposed framework using the \emph{mprof} tool. The memory profile
of the proposed framework for two benchmark circuits $ibmpg2$ and $ibmpg6$ are shown in Figure \ref{fig_peak_memeory}.
We also show the peak memory usage for all the IBM PG benchmarks as listed in Table \ref{table_mse}. 

\section{Conclusion and Future Work}\label{sec_con}
In this paper, we have proposed a deep learning-based framework PowerPlanningDL to predict the initial power grid design.
For the first time, we have shown the equivalence between the neural network training and power grid design.
We predict the power grid interconnect width as part of the design process, which is time-consuming and tedious work.
Subsequently, we also anticipate the worst-case IR drop in the power grid. A neural network-based multi-regression technique is used in our model for accomplishing the prediction tasks. Results on IBM power grid benchmarks show $\sim$6$\times$ speedup
than the conventional power grid design approach. We have also performed various other experiments.

From the results of the experiments, we can recommend the following for the adaptation of the deep learning in power planning phase of VLSI Physical Design:
\begin{itemize}
 \item The predictability of the deep learning approach is close to the conventional method, with very less convergence time ($\sim$6$\times$ speedup).
 \item Deep learning in power planning is useful in the incremental-based power grid designs, where the perturbation size is small.
 \item The error due to the prediction increases for the PowerPlanningDL framework for the designs with large perturbations.
 \item Finally, from this work, we can say that the industry can adapt the deep learning approach for the power
 grid design, which will reduce many iterative steps in order to obtain an appropriate initial design.
\end{itemize}
Further, a better learning approach can be introduced for the efficient power grid design. Additionally, decap placement-aware power grid design using deep learning technique can also be explored. 
\balance

%\setstretch{1.5}

%\vspace{-0.2cm}
{\tiny
\bibliographystyle{IEEEtran}
\bibliography{ref}}

\end{document}